\pdfoutput=1

\documentclass[11pt]{article}

\usepackage[]{emnlp2021}

\usepackage{times}
\usepackage{latexsym}

\usepackage[T1]{fontenc}

\usepackage[utf8]{inputenc}

\usepackage{microtype}

\usepackage{url}
\usepackage[utf8]{inputenc}
\inputencoding{utf8}
\usepackage{CJKutf8}
\usepackage{multirow}
\usepackage{lipsum}
\usepackage{url}
\usepackage{amsmath}
\usepackage{bm}
\usepackage{todonotes}
\usepackage{wrapfig}
\usepackage{graphicx}
\usepackage{float}
\usepackage{tikz}
\usetikzlibrary{automata,positioning,arrows}
\usetikzlibrary{decorations.text}
\usepackage{bbm}
\usepackage{threeparttable}

\usepackage{caption}
\usepackage{latexsym}
\usepackage{amssymb}
\usepackage{amsmath}
\usepackage{subcaption}
\usepackage{algorithm,algpseudocode}
\usepackage{footnote}
\makesavenoteenv{tabular}
\usepackage[export]{adjustbox}

\usepackage{siunitx} 
\usepackage{booktabs}
\title{Back-translation for Large-Scale Multilingual Machine Translation}

 \author{
  Baohao Liao \qquad Shahram Khadivi \qquad Sanjika Hewavitharana \\
  eBay Inc. \\
  \texttt{\{baliao|skhadivi|shewavitharana\}@ebay.com}
  }

\begin{document}
\maketitle

\begin{abstract}
This paper illustrates our approach to the shared task on large-scale multilingual machine translation in the sixth conference on machine translation (WMT-21). This work aims to build a single multilingual translation system with a hypothesis that a universal cross-language representation leads to better multilingual translation performance. We extend the exploration of different back-translation methods from bilingual translation to multilingual translation. Better performance is obtained by the constrained sampling method, which is different from the finding of the bilingual translation. Besides, we also explore the effect of vocabularies and the amount of synthetic data. Surprisingly, the smaller size of vocabularies perform better, and the extensive monolingual English data offers a modest improvement. We submitted to both the small tasks and achieved the second place. The code and trained models are available at \url{https://github.com/BaohaoLiao/multiback}.
\end{abstract}

\section{Introduction}
Bilingual neural machine translation (NMT) systems have achieved decent performance with the help of Transformer \cite{vaswani2017attention}. One of the most exciting recent trends in NMT is training a single system on multiple languages at once \cite{10.1162/tacl_a_00065, aharoni-etal-2019-massively, zhang-etal-2020-improving, DBLP:journals/corr/abs-2010-11125}. This is a powerful paradigm for two reasons: simplifying system development and deployment and improving the translation quality on low-resource language pairs by transferring similar knowledge from high-resource languages.

This paper describes our experiments on the task of large-scale multilingual machine translation in WMT-21. We primarily focus on the small tasks, especially on Small Task 2, which has a small amount of training data. Small Task 1 contains five Central/East European languages and English, having 30 translation directions. Similarly, Small Task 2 contains five South East Asian languages and English, also having 30 translation directions. 

In this work, we mainly concentrate on different back-translation methods \cite{sennrich2016aimproving, edunov2018understanding, gracca2019generalizing} for multilingual machine translation, including beam search and other sampling methods. Along with it, we also explore the effect of different sizes of vocabularies and the effect of various amounts of synthetic data. On this large-scale multilingual machine translation task, we achieved the second place for both small tasks, obtaining 34.96 and 33.34 average spBLEU scores \cite{goyal2021flores} on the hidden test set for Small Task 1 and 2, respectively.

\section{Related Work}
\textbf{Multilingual Neural Machine Translation}  has received increasing attention recently. Since \citet{dong-etal-2015-multi} extended the traditional bilingual NMT to
one-to-many translation, there has been a massive increase in work on MT systems that involve more than two languages \cite{DBLP:journals/corr/DabreCK17, CHOI18.139, DBLP:journals/corr/abs-1906-07978}. The recent research on multilingual NMT can be split into two directions: developing language-specific components \cite{kim-etal-2019-effective, DBLP:journals/corr/abs-2006-01594} and training a single model with extensive training data, including parallel and monolingual data \cite{DBLP:journals/corr/abs-2010-11125}. Here, we continue to explore the second research direction, trying to build a single multilingual NMT model for simple industrial deployment.

\textbf{Back-translation} \cite{sennrich2016aimproving} has been proven as a powerful technique to leverage monolingual data for improving low-resource language pairs. \citet{edunov2018understanding} and \citet{gracca2019generalizing} explore different sampling methods for bilingual back-translation, including beam search, constrained and unconstrained sampling. Constrained sampling randomly predicts the next word within some candidates that have a higher prediction probability. And unconstrained sampling randomly predicts the next words from the whole vocabulary without caring for the output distribution. In this paper, we extend their exploration to the realm of multilingualism, where similar languages affect the results.

\section{ Experimental Setup}

\subsection{Data}
The organizer offers parallel and monolingual data for Small Task 1 and 2. Table \ref{tab:data amount} shows the size of the data in terms of the number of sentences for each language. There are five extra sets for evaluation, i.e., dev, devtest, hidden dev, hidden devtest, and test sets. The dev set with 997 parallel sentences among all language pairs and the devtest set with 1,012 parallel sentences are public. In contrast, the hidden dev and hidden devtest sets are invisible to the participants and used for the first submission period. The hidden test set is also invisible and used for the final ranking.

Pre-processing is done by a regular Moses toolkit \cite{koehn2007moses} pipeline that involves tokenization, byte pair encoding and removing long sentences. We borrow the 256K vocabularies from the organizer's pretrained model and the 128K vocabularies from M2M\_100 \cite{fan2021beyond}, one shared vocabulary among all languages. Our submissions only use the 256K vocabularies, while the 128K vocabularies is used for ablation experiments.

We also perform back-translation on the monolingual data and only accept the synthetic sentence pair whose length is less than 250 words and whose length ratio between the source and target sentence length is less than 1.8. In order to balance the volume across different languages, we apply temperature sampling $\tilde{D}_i = (D_i / \sum_j{D_j})^{1/T}$ with $T=5$ over the dataset, where $D_i$ is the number of sentences in the $i_{th}$ language.

\begin{table}[t]
  \begin{center}
    \begin{tabular}{lr|lr} 
      \toprule
      \multicolumn{2}{c|}{\textbf{Small Task 1}} & \multicolumn{2}{c}{\textbf{Small Task 2}} \\
      \textbf{Language} & \textbf{\#sent.} & \textbf{Language} & \textbf{\#sent.}  \\
      \midrule
      en-et & 35.7M &  en-id & 54.1M \\
      en-hr & 63.7M & en-jv & 3.0M \\
      en-hu & 83.9M & en-ms & 13.4M \\
      en-mk & 2.7M & en-ta & 2.1M \\
      en-sr & 48.3M & en-tl & 13.6M \\
      et-hr & 13.6M & id-jv & 780.1K \\
      et-hu & 21.5M & id-ms & 4.9M \\
      et-mk & 3.1M & id-ta & 500.8K \\
      et-sr & 11.3M & id-tl & 2.7M \\
      hr-hu & 31.2M & jv-ms & 434.7K  \\
      hr-mk & 4.4M & jv-ta & 66.0K \\
      hr-sr & 28.4M & jv-tl & 817.1K \\
      hu-mk & 4.1M & ms-ta & 372.6K \\
      hu-sr & 31.2M & ms-tl & 1.4M \\
      mk-sr & 4.2M & ta-tl & 563.3K \\
      \hline 
      en & 126.4M & en & 126.4M \\
      et & 3.0M & id & 5.5M \\
      hr & 3.1M & jv & 405.8K \\
      hu & 9.2M & ms & 1.9M \\
      mk & 1.9M & ta & 2.1M \\
      sr & 4.7M & tl & 414.1K \\
      \bottomrule
    \end{tabular}
    \caption{Number of sentences of the parallel and monolingual data used for two small tasks. The monolingual English data for the two small tasks are the same.}
    \label{tab:data amount}
  \end{center}
\end{table}

\begin{table*}[t]
  \begin{center}
    \begin{tabular}{cccc} 
      \toprule
      \textbf{Model} & \textbf{Trans\_small} & \textbf{Trans\_base} & \textbf{Trans\_big} \\
      \midrule
      \#vocabularies &  256K & 256K & 128K \\
      Word representation size & 512 & 1,024 & 1,024 \\
      Feed-forward layer dimension & 2,048 & 4,096 & 8,192 \\
      \#prenormed encoder/ decoder layer & 6 & 12 & 24 \\
      \#attention head &  16 & 16 & 16 \\
      Dropout rate & 0.1 & 0.1 & 0.1 \\
      Layer dropout rate & 0.05 & 0.05 & 0.05 \\
      \#parameters & 175M & 615M & 1.2B \\
      \bottomrule
    \end{tabular}
    \caption{Settings of different pretrained models. Pretrained \textit{Trans\_small} and \textit{Trans\_base} are provided by the organizer. And pretrained \textit{Trans\_big} is from \citet{fan2021beyond}.}
    \label{tab:model settings}
  \end{center}
\end{table*}
\subsection{Model}
All our models are built using the fairseq implementation \cite{ott2019fairseq} of the Transformer architecture \cite{vaswani2017attention}. Multilingual models are
built using the same technique as \citet{johnson2017google} and \citet{aharoni2019massively}, namely adding a language label to the target sentence.

We apply three types of architectures, i.e., \textit{Trans\_small}, \textit{Trans\_base}, and \textit{Trans\_big}. The detailed settings of these architectures are shown in Table \ref{tab:model settings}. The parameters of all architectures are in the half-precision floating-point format.

All our submissions on the shared task leaderboard are \textit{Trans\_base} due to the memory and time limit of the evaluation system. \textit{Trans\_small} is mainly used for the ablation experiments. And the pretrained \textit{Trans\_big} from M2M\_100 \cite{fan2021beyond} is finetuned on the parallel corpus to generate high-quality synthetic sentences. 

\subsection{Optimization and Evaluation}
The following hyper-parameter configuration is used: Adam optimizer with $\beta_1 = 0.90$, $\beta_2 = 0.98$, a weight-decay of 0.0001, the label smoothed cross-entropy criterion with a label smoothing of 0.1, an initial learning rate of 0.0003 with the inverse square root lr-scheduler and warmup updates of 2,500 steps. The batch size (the number of tokens) is $4096 \times 32$ for \textit{Trans\_small}, and $2048 \times 64$ for \textit{Trans\_base} and \textit{Trans\_big}. 

For ablation experiments, we continue to train the pretrained \textit{Trans\_small} offered by the organizer on the given parallel dataset for one epoch. We further train the model finetuned on the parallel data for another epoch when combining both parallel and synthetic data. For the final submissions, we train a pretrained \textit{Trans\_base} for two epochs instead of one epoch. Pretrained \textit{Trans\_big} from M2M\_100 is only further trained on parallel data for two epochs to generate high-quality synthetic data. Even though we only train these models for a few epochs, they seem to converge quite well according to the spBLEU curve during validation.

The model is validated every 3,000 steps on the dev set and saved. We use the beam search with a beam size of five, and stop translation when $l_{tgt} = 1.5 * l_{src} + 20$, where $l_{src}$ and $l_{tgt}$ are the source and target sentence length, respectively. The evaluation metric is BLEU based on sentence piece tokenization (spBLEU) \cite{goyal2021flores}. We submit the average checkpoint of the last 15 checkpoints to the evaluation system. While for the ablation experiment, we use the best-performed model on the dev set.

\section{Results}
\subsection{The Role of Vocabularies} \label{sec:The Role of Vocabularies}
There are two pretrained vocabularies, the one with the size of 256K from the organizer and the one with the size of 128K from M2M\_100 \cite{fan2021beyond}. To evaluate which vocabulary is better, we train two \textit{Trans\_smalls} with these two vocabularies from scratch on the parallel data of Small Task 2 for five epochs. To make the parameter sizes of these two models comparable, we set the following hyper-parameter for the model with the 128K vocabularies: 5 pre-normed encoder and decoder layers with a word representation size of 768 and a feed-forward layer dimension of 3072, resulting in 181M parameters. The other settings stay the same with \textit{Trans\_small} (with 256K vocabularies).

Table \ref{tab:voc} shows the performance with different vocabularies. It is obvious that the 128K vocabulary outperforms the 256K vocabulary, 23.14 vs. 21.65 spBLEU. However, if we finetune the pretrained \textit{Trans\_small} with the 256K vocabulary, 0.58 score improvement is achieved compared to the 128K \textit{Trans\_small}. In a word, 128K vocabulary is a better choice for training from scratch, while pretrained model offers us more gain.

\begin{table}[t]
  \begin{center}
    \begin{tabular}{lc} 
      \toprule
      \textbf{Model} & \textbf{Ave. spBLEU} \\
      \midrule
      128K \textit{Trans\_small} (scratch) & 23.14  \\
      256K \textit{Trans\_small} (scratch) & 21.65 \\
      256K \textit{Trans\_small} (pretrained) & 23.72 \\
      \bottomrule
    \end{tabular}
    \caption{Average spBLEU on the devtest set of Small Task 2 for the models with different vocabularies.}
    \label{tab:voc}
  \end{center}
\end{table}

\begin{table}[t]
  \begin{center}
    \begin{tabular}{lc} 
      \toprule
      \textbf{Model} & \textbf{Ave. spBLEU} \\
      \midrule
      1st finetuned on parallel data & 28.27  \\
      2nd finetuned on synthetic data & 32.16 \\
      3rd finetuned on synthetic data & 33.01 \\
      \bottomrule
    \end{tabular}
    \caption{Average spBLEU on the devtest set of Small Task 2 for \textit{Trans\_base} on different finetuning steps. These three models are iteratively trained. \textit{Trans\_base} is first finetuned on the parallel data, and then finetuned on the combination of the parallel data and the synthetic data generated by \textit{Trans\_big}, and finally finetuned on the combination of the parallel data and the synthetic data generated by the 2nd step \textit{Trans\_base}.}
    \label{tab:iterative finetuning}
  \end{center}
\end{table}

\begin{table}[t]
  \begin{center}
    \begin{tabular}{lc} 
      \toprule
      \textbf{Model} & \textbf{Ave. spBLEU} \\
      \midrule
      1st finetuned on parallel data & 32.46  \\
      2nd finetuned on synthetic data & 34.73 \\
      \bottomrule
    \end{tabular}
    \caption{Average spBLEU on the devtest set of Small Task 1 for \textit{Trans\_base} on different steps. These two models are iteratively trained. \textit{Trans\_base} is first finetuned on the parallel data, and then finetuned on the combination of the parallel data and the synthetic data generated by the previous step \textit{Trans\_base}.}
    \label{tab:iterative finetuning task1}
  \end{center}
\end{table}

\begin{table*}[t]
  \begin{center}
    \begin{tabular}{ccccc} 
      \toprule
      \textbf{Small Task} & \textbf{devtest} & \textbf{hidden dev} & \textbf{hidden devtest} & \textbf{hidden test} \\
      \midrule
      \#1  & 34.73 & 35.12 & 35.39 & 34.96 \\
      \#2  & 33.01 & 33.74 & 33.51 & 33.34 \\
      \bottomrule
    \end{tabular}
    \caption{Average spBLEU on different test sets for both small tasks. The hidden sets are invisible to the participants. The final ranking is based on the model performance on the hidden test set.}
    \label{tab: finla submission}
  \end{center}
\end{table*}

\begin{figure}[t]
  \centering
    \includegraphics[width=0.5\textwidth]{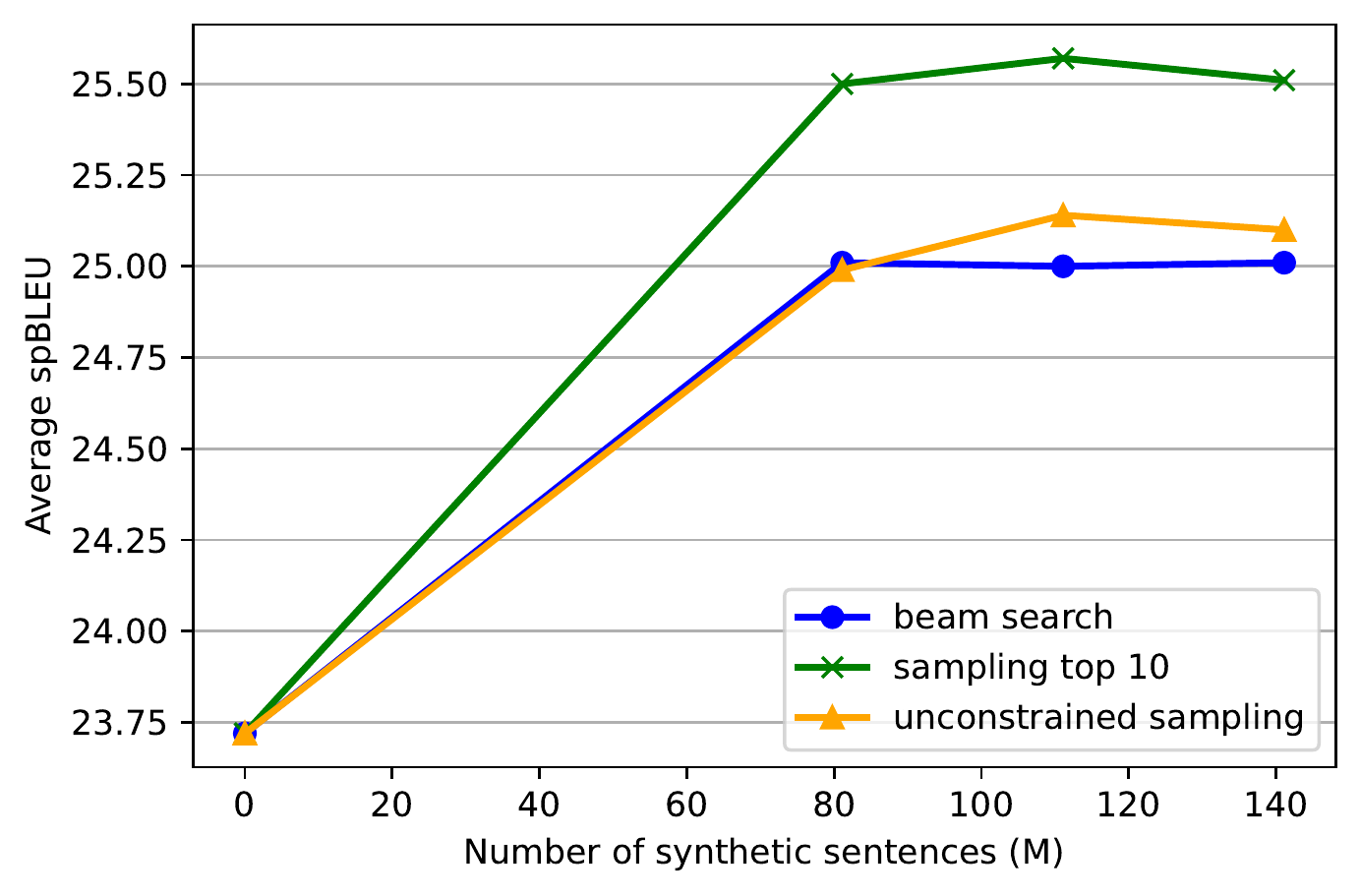}
    \caption{Average spBLEU on the devtest set of Small Task 2 for different back-translation methods with various amounts of synthetic data. 80M synthetic data covers only 6M monolingual English data and all other monolingual data. We increase the amount of monolingual English data with an interval of 6M for the last two experiments.}
    \label{fig:backtranslation}
\end{figure}

\begin{figure*}[t]
  \begin{subfigure}[b]{0.5\textwidth}
        \centering
        \includegraphics[width=1\linewidth]{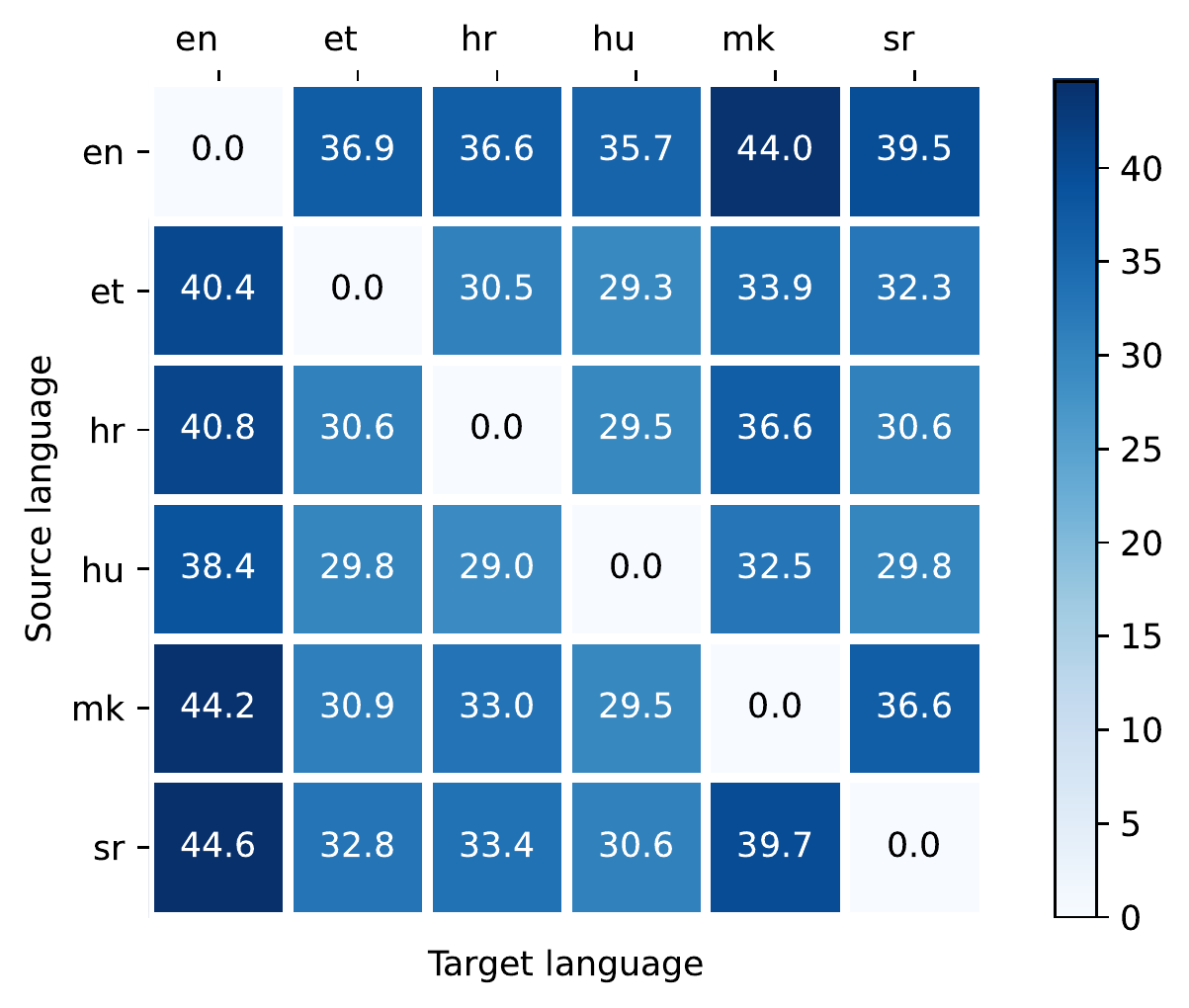}
  \end{subfigure}
  \begin{subfigure}[b]{0.5\textwidth}
        \centering
        \includegraphics[width=1\linewidth]{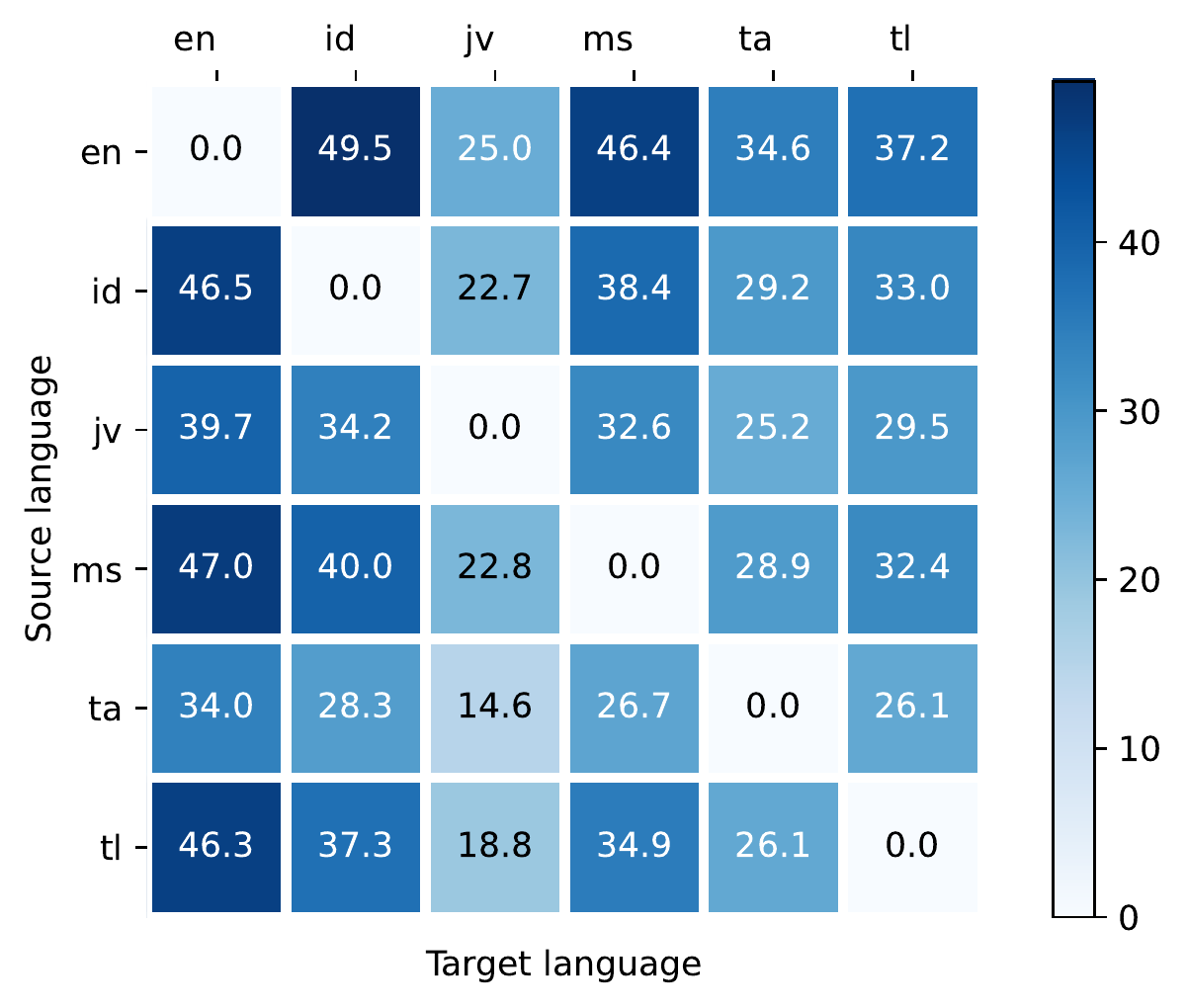}
  \end{subfigure}
\caption{The spBLEU scores of different language pairs for both small tasks on the devtest set from our final submissions.}
\label{fig:heatmap}
\end{figure*}

\subsection{Different Back-translation Methods} \label{sec:Different Back-translation Methods}
Similar to \citet{edunov2018understanding}, we explore three types of back-translation methods, i.e., beam search with the beam size of five \cite{sennrich2016aimproving}, unconstrained sampling \cite{edunov2018understanding} and sampling constrained to the most 10 likely words \cite{graves2013generating, ott2018analyzing, fan2018hierarchical}. Unconstrained sampling predicts the next word from the whole vocabulary without caring for the model distribution. In contrast, constrained sampling predicts the next words within some candidates that have the highest prediction probabilities. Both constrained and unconstrained sampling can be considered as adding uncertainty to the greedy search.  

Figure \ref{fig:backtranslation} shows the back-translation results on the devtest set of Small Task 2. We combine three different amounts of synthetic data and parallel data to further train our \textit{Trans\_small}s after finetuned on parallel data. 80M synthetic sentences cover only 6M monolingual English data and all other monolingual data. In addition to the 80M synthetic sentences, we further increase the amount of monolingual English data to verify the model performance concerning the amount of synthetic English data on the target side. The reason for this implementation is there are too many monolingual English sentences compared to other languages. We try to check whether it is necessary to use all monolingual English sentences.

As seen in Figure \ref{fig:backtranslation}, little improvement is obtained by increasing monolingual English sentences after 6M. Besides, in contrast to the results in \citet{edunov2018understanding} where the unconstrained sampling offers the best performance among these three methods, the constrained sampling method gives us the best score.

Beam search is the worst among these three methods. We hypothesize this is because beam search focuses only on the high probability words, while both constrained sampling and unconstrained sampling methods offer rich translations on the source side. With the diverse synthetic data generated from the sampling methods, the model can be trained with a higher level of generalization. 

In contrast to the bilingual translation (English-German) in \citet{edunov2018understanding} where unconstrained sampling outperforms constrained sampling, multilingual translation of Small Task 2 contains similar languages. We argue that unconstrained sampling might generate synthetic sentences with a mix of similar languages, which damages the quality of synthetic data, while constrained sampling gives us some restriction, to some extent avoiding the mix of different languages. 

The reason for the slight effect of the synthetic English (on the source side) data after 6M might be that English is dissimilar to the other five South East Asian languages. Less similar knowledge could be transferred from this synthetic English (on the source side) data to other languages.

\subsection{Final Submissions}
Section \ref{sec:The Role of Vocabularies} suggests us to employ a pretrained model with the 128K vocabulary. M2M\_100 \cite{fan2021beyond} offers multiple pretrained models with the 128K vocabularies \footnote{\url{https://github.com/pytorch/fairseq/tree/master/examples/m2m_100}}. Their sizes are 418M, 1.2B, and 12B, respectively. Considering our limited GPU budget, we finetune the 1.2B model, i.e. \textit{Trans\_big}, on parallel data of Small Task 2, obtaining 28.78 spBLEU on the devtest set. In comparison, training a \textit{Trans\_base} on the same data only provides 28.23 spBLEU. Even though \textit{Trans\_big} outperforms \textit{Trans\_base}, we only train it for generating high-quality synthetic data since it is too large for the evaluation system.

Section \ref{sec:Different Back-translation Methods} advises us to use the constrained sampling method on partial monolingual English data. With the constrained sampling method, we generate synthetic sentences with \textit{Trans\_big} that is first finetuned on the parallel data. Instead of using all monolingual English data, we synthesize en-id, en-jv, en-ms, en-ta and en-tl with all, 15M, 60M, 10M, and 60M monolingual English sentences, respectively, a ratio of about $5:1$ between the number of parallel sentences and synthetic sentences if there are enough monolingual data. 

Table \ref{tab:iterative finetuning} shows the results for iterative finetuning. Except for finetuning \textit{Trans\_base} on the combination of the parallel data and the synthetic data generated by \textit{Trans\_big}, we use the finetuned \textit{Trans\_base} to generate the synthetic data secondly and finetune it again. Finally, it offers us 33.01 spBLEU on the devtest set for Small Task 2. 

Due to time and resource limits, we only conduct one trial on Small Task 1. We first finetune the pretrained \textit{Trans\_base} on parallel data. Then we use this \textit{Trans\_base} to generate synthetic data with only 20M monolingual English sentences and all other monolingual sentences. Table \ref{tab:iterative finetuning task1} shows the corresponding results. Different from Small Task 2, a large amount of monolingual English data might be helpful for Small Task 1 since Central/East European languages are more similar to English than Asian languages. Finally, We leave this exploration to future work.

Table \ref{tab: finla submission} summarizes the results of our submissions on different evaluation sets for both small tasks. And Figure \ref{fig:heatmap} lists the spBLEU scores for all language pairs of both small tasks on the devtest set. Finally, our submissions achieved the second place for both small tasks.

\section{Conclusion}
We demonstrate that a pretrained model with a smaller size of the vocabulary is a better choice. Because of the memory and time limit of the evaluation system, we can only apply a 1.2B model with the smaller vocabularies to generate high-quality synthetic data. Besides, we have a different observation than previous research for bilingual back-translation: the constrained sampling method performs the best among all three back-translation methods, including the beam search and the unconstrained sampling. Finally, we also show that extensive monolingual English data offers a modest improvement. Combining these three findings, we iteratively train our models on partial high-quality synthetic data, achieving the second place for both small tasks. 

\bibliography{anthology,custom}
\bibliographystyle{acl_natbib}

\clearpage
\appendix


\end{document}